\documentclass{article}

\usepackage{arxiv}

\usepackage[utf8]{inputenc} 
\usepackage[T1]{fontenc}    
\usepackage{hyperref}       
\usepackage{url}            
\usepackage{booktabs}       
\usepackage{amsfonts}       
\usepackage{nicefrac}       
\usepackage{microtype}      
\usepackage{lipsum}
\usepackage{graphicx}
\usepackage{cite}
\usepackage{amsmath,amssymb,amsfonts}
\usepackage{graphicx}
\usepackage{textcomp}
\usepackage{xcolor}
\usepackage{mathtools, cuted}
\usepackage{multirow}
\usepackage{color,soul}
\usepackage{xspace}
\usepackage{subfig}
\usepackage{array}
\usepackage{algorithm}
\usepackage{algpseudocode}
\usepackage{hyperref}
\usepackage[toc,page]{appendix}
\usepackage{adjustbox}
\graphicspath{ {./images/} }

\title{Towards optimized actions in critical situations of soccer games with deep reinforcement learning}
\date{}

\author{
 Pegah Rahimian \\
  Budapest University of Technology\\ and Economics\\
  Budapest, Hungary \\
  \texttt{pegah.rahimian@tmit.bme.hu} \\
   \And
 Afshin Oroojlooy \\
  SAS Institute\\
  Cary, NC, USA \\
  \texttt{afshin.oroojlooy@sas.com} \\
  \And
 Laszlo Toka \\
  MTA-BME Information Systems\\ Research Group\\
  Budapest, Hungary\\
  \texttt{toka.laszlo@vik.bme.hu} \\
}

\begin{document}
\maketitle
\begin{abstract}
Soccer is a sparse rewarding game: any smart or careless action in critical situations can change the result of the match. Therefore players, coaches, and scouts are all curious about the best action to be performed in critical situations, such as the times with a high probability of losing ball possession or scoring a goal. This work proposes a new state representation for the soccer game and a batch reinforcement learning to train a smart policy network. This network gets the contextual information of the situation and proposes the optimal action to maximize the expected goal for the team. We performed extensive numerical experiments on the soccer logs made by InStat for 104 European soccer matches. The results show that in all 104 games, the optimized policy obtains higher rewards than its counterpart in the behavior policy. Besides, our framework learns policies that are close to the expected behavior in the real world. For instance, in the optimized policy, we observe that some actions such as foul, or ball out can be sometimes more rewarding than a shot in specific situations.
\end{abstract}


\section{Introduction}

Soccer is a complex and sparse game, with a large variety of actions, outcomes, and strategies. These aspects of a soccer game make the analysis strenuous. Recent breakthroughs in computer vision methods help sport analysis companies, such as InStat\cite{instat}, Wyscout\cite{wyscout}, StatsBomb\cite{statsbombs}, STATS\cite{stats}, Opta\cite{opta}, etc., collect highly accurate tracking and event datasets from match videos. 
Obviously, the existing tracking and event data in the market contain the prior decisions and the observed outcomes of players and coaches, with some nonrandom, regular, and non-optimized policies. We call these policies as behavioral policies throughout the rest of this paper.
Nowadays, analyzing the behavioral policies obeyed by the players and dictated by coaches, has been one of the most interesting topics for researchers. Sports analysts, i.e., academic researchers, applications, scouts, and other sports professionals, are investigating the potential of using previously collected, i.e., off-line, data to make counterfactual inference of how alternative decision policies could perform in a real match. 

There are several engrossing action valuation methods in the literature of sports analytics, focusing on passes and shots (e.g., \cite{Fernandez2019, Fernandez2020, fernandez2021, gyarmati2016qpass}, etc.), and some others cover all type of actions (e.g., \cite{Tom2019, Liu2018, Liu2019}, etc.). They accurately evaluate the player actions and contribution to goal scoring. However, all those models leave the player and the coach with the value of the performed action, without any proper proposal of alternative and optimal actions. 
To fill this gap, this work goes beyond action valuation by proposing a novel policy optimization method, which can decide about the optimal action to perform in critical situations. In soccer, we consider as critical situations the moments with a high probability of losing the ball, or scoring/conceding a goal. However, the player does not have any chance of passing to teammates, or dribbling in these situations. Thus, she/he needs to immediately decide about the following options: 1) shooting, 2) sending the ball out, 3) committing a foul, 4) submitting the ball to the opponent by making an error. Moreover, we define the optimal action as the action that maximizes the expected goal for the team. Thus, our method should both evaluate the behavioral policy, and suggest the optimal target policy to the players and coaches, for critical situations. It is a challenging task to design such a system in soccer due to the following reasons: first, soccer is a highly interactive and sparse rewarding game with a relatively large number of agents (i.e., players). Thus, state representation is ambiguous in such a system and requires an exact definition. Second, the spatiotemporal and sequential nature of soccer events and dynamic players' location on the field dramatically increases the state dimensions, which is never pleasant for machine learning tasks. Third, the game context in a soccer match is severely affecting the model prediction performance. Forth, evaluating a trained optimal policy requires deployment in a real soccer match. However, this solution sounds impossible due to the large cost of deployment. This work offers solutions to all the above-mentioned challenges. Sports professionals can use our policy optimization method after the match to check what action the player performed, evaluate it, and propose the optimal alternative action in that critical situation. If the action performed by the player, and the optimal action proposed by the optimal policy are not the same, we can relate it to the player's mistake, or poor strategy from the coach.

In summary, our work contains the following contributions:

\begin{itemize}
\item We propose an end-to-end framework that utilizes raw data and trains an optimal policy to maximize the expected goal of soccer teams via reinforcement learning (RL);
\item Introduce a soccer ball possession model, which we assume to be Markovian, and a new state representation to analyze the impact of actions on subsequent possessions;
\item Suggest spectral clustering with regards to the opponent's position and velocity for measuring the pressure on the ball holder at any moment of the match;
\item Propose a new reward function for each time-step of the game, based on the output of the neural network predictor model;
\item Derive the optimal policy in critical situations of soccer matches, with the help of fully off-policy, deep reinforcement learning method.
\end{itemize}

\section{Related work}
\label{sec:sot}

The state-of-the-art models in soccer analytics are focusing on several aspects such as evaluating actions, players, and the strategies.    
Plus/minus method is an early work on player evaluation that has been proposed by Kharrat et al. \cite{Kharrat2017}. This method assigns plus for each goal scored and minus for each goal conceded by the players per total time they were on the pitch. Although this is the simplest method, it ignores the rating of other players, the opposition strength, and does not account for match situations. 
Regression method on actions and shots was firstly proposed by Ian et al. \cite{Ian2012}. They estimate the number of shots as the function of crosses, dribbles, passes, clearances, etc. Coefficients show how important they are in generating shots. However, this model does not work well in some cases (e.g., when the value of pass changes, and in case we want to know where the cross occurred). 
Another interesting player evaluation method is percentiles and player radars by Statsbomb \cite{Statsbomb}. This method estimates the relative rank for each player based on his actions. For example, a ranking can be assigned to a player for all his defensive actions (tackle/interception/block), his accurate passes, crosses, etc. 

The application of a Markovian model in action valuation was first proposed by Rudd \cite{Rudd,GoldnerKeith2012}. The input of this model is the probability of ball location in the next five seconds. Assuming we have these probabilities, this model estimates the likely outcomes after many iterations based on the probabilities of transitioning from one state to another. 
Another application of the Markovian model is the Expected threat (xT) \cite{xT}, which uses simulations of soccer matches to assign value to the actions. Although, we believe that simulations tend to be unrealistic. Because the simulations with any arbitrary point are not resulting in a goal by several iterations. 
VAEP \cite{Tom2019} is another action valuation model, which considers all types of actions. This model uses a classifier to estimate the probability that an action leads to a goal within the next 10 actions, and the game state is considered as 3 actions. This model ignores the concept of possessions in valuation. 
Considering the possession, the Expected Possession Value (EPV) metrics in football \cite{Fernandez2019} and basketball \cite{cervone2014} were proposed. These models assume a simple world in which the actions of the players inside possessions are limited to pass, shot, and dribble. Thus, ignoring any other actions such as foul, ball out, or the errors, which frequently happen in critical situations.

Recently, researchers utilize deep learning methods due to their promising performance in valuation domains. 
Fernandez and Bornn \cite{fernandez2021} present a convolutional neural network architecture that is capable of estimating full probability surfaces of potential passes in soccer. 
Moreover, Liu et al. took advantage of RL, by assigning value to each of the actions in ice-hockey \cite{Liu2018} and soccer \cite{Liu2019} using Q-function. They later used a linear model tree to mimic the output of the original deep learning model to solve the trade-off between Accuracy and Transparency \cite{Xiangyu2020}. Moreover, Dick and Brefeld \cite{Dick2019} used reinforcement learning to rate player positioning in soccer.
In this paper, we go beyond the valuation of actions in critical situations, and use RL to derive the optimal policy to be performed by the teams and players.

\section{Our Markovian possession model}
\label{sec:markov}

In order to train an RL model, we first represent a soccer game as a Markov decision process. To this end, in this section we introduce an episode of the game, the start, intermediate, and final states. In the next sections, we define the state, action, and reward in each time-step of the game. 


Due to the fluid nature of a soccer game, it is not straightforward to have a comprehensive description of a possession, which applies to all different types of soccer logs provided by different companies (e.g., InStat, Wyscout, StatsBomb, Opta, etc.). In the InStat dataset and accordingly in this work, possessions for any home or away teams are clearly defined and numbered. A possession starts from the beginning of a deliberate and on the ball action by a team, until it either ``ends'' due to some event like ball out, foul, bad ball control, offside, clearance, goal (regardless of who possesses the ball afterward, i.e., the next possession can belong to the same team or opposing team), or ``transfers'' by a defensive action of the opponent, such as pass interception, tackle, or clearance. 

The possession can be transferred if and only if the team is not in the possession of the ball over two consecutive events. Thus, the unsuccessful touches of the opponent in fewer than 3 consecutive actions are not considered as a possession loss. Consequently, all on-the-ball actions of players of the same team should be counted to get the possession length, not only passes, shots, and dribbles. Accordingly, we define an episode $\tau$ as subsequent possessions for any team, until they lose the ball, or end possession sequences with a shot.

We aspire to describe the possessions with a Markovian model.
In order to take advantage of the Markovian model of possessions and their outcomes, we converted the action level nature of the dataset to possession level, and each possession is labeled by its own terminating action. This conversion expedites the usage of supervised learning methods to predict the most probable outcomes as well. Our proposed model can be separately applied to any team participating in the games. 

\begin{figure}[h]
    \centering
  \includegraphics[ width=6cm]{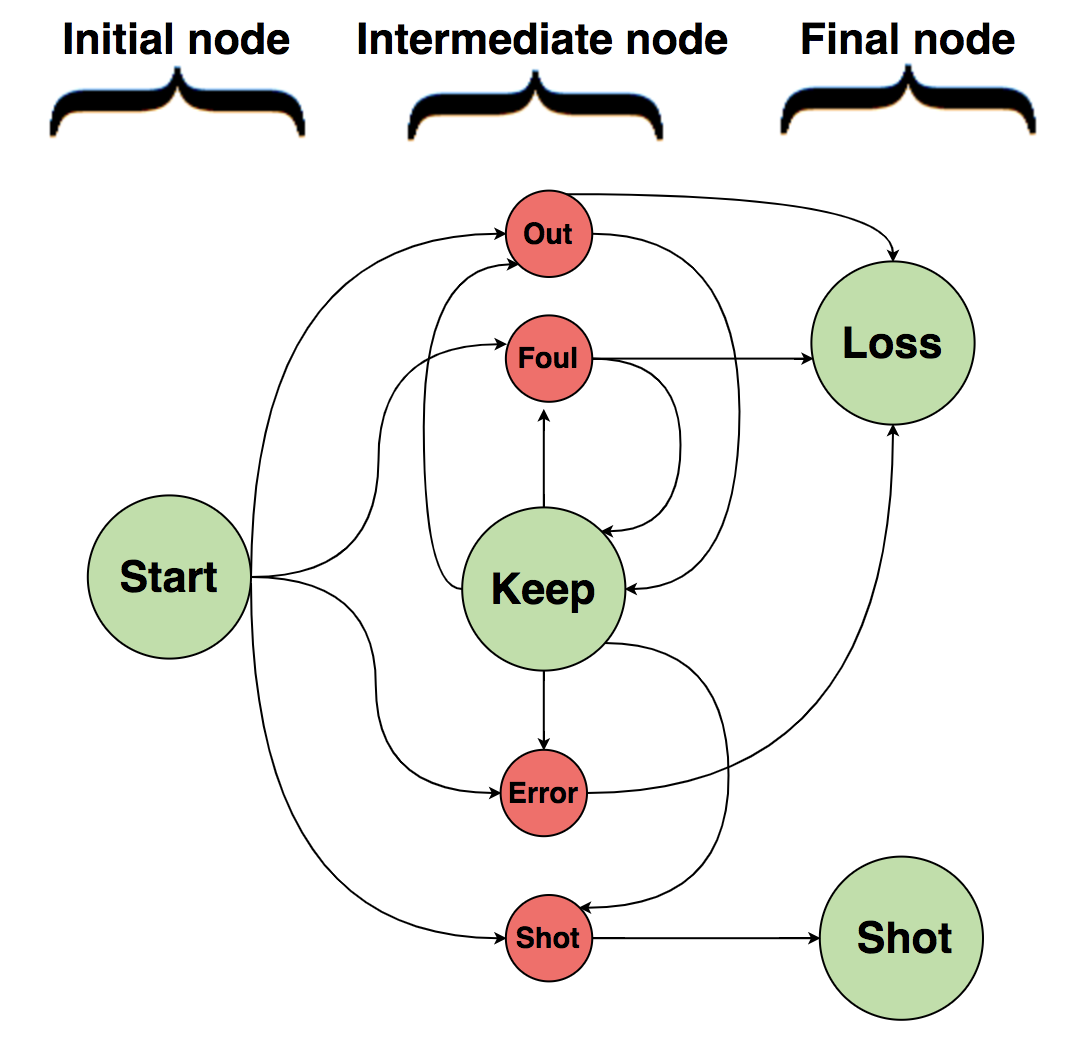}
  \caption{
  Finite State Automaton of the Markovian possession model. The state is considered as one possession. Green nodes show the conditions of the possessions, transited by the ending actions. Red circles are actions categorized by intentional (out, foul, shot) or unintentional (errors, i.e., players mistakes such as bad ball control, pass inaccurate, or tackles by opponents).}
  \label{fsa}
\end{figure}

We can model this process as a Finite State Automaton, with the initial node of ``Start'' of possession, the final nodes of (``Loss'' or ``Shot''), and the intermediate node of ``Keep'' the possession. The schematic view of the state transition is illustrated in Figure~\ref{fsa}. 




\section{State representation and neural network architecture}\label{sec:dataset_state_generation_cnn}
In this study, we present an end-to-end framework to learn an optimal policy for maximizing the expected goals in a soccer game. To achieve this goal, data preparation is a core task to achieve a reliable RL model. In this section, we present the steps of building the states. Considering the definition of the episode in Section~\ref{sec:markov}, there is no necessity for existence of a goal in an episode; so, we need to define a well-suited reward function for each time-step. We propose a neural network model, utilizing the suggested state, and obtain the underlying data to get the reward of each time-step. The structure of the datasets used in this study is provided in Appendix~\ref{data}.

\subsection{Game context: opponent pressure}
\label{clustering-sec}

Considering a descriptive game context is one of the most important aspects of soccer analytics, when it comes to feature engineering.
Several works introduced different methods, KPIs, and features to address this problem. Among the works, Decroos et al. \cite{Tom2019} created the following game context features: number of goals scored by attacking team after action, number of goals scored by defending team after action, and goal difference after action. 
Fernandez et al. \cite{Fernandez2019} considered the role of context by slicing the possession into 3 phases: build-up, progression, and finalization. They considered three dynamic formation or pressure lines, and grouped the actions based on particular relative locations: first vertical pressure line (forwards), second vertical pressure line (midfielders), third vertical pressure line (defenders). 
Another interesting approach by Alguacil et al. \cite{Fernandez2020} mimicked the collective motion of animal groups, called self-propelled particles, in soccer. They claimed that in FC Barcelona, and generally, coaches can talk about three different playing zones: intervention zone (immediate points around the ball), mutual help zone (players close to the ball, but further away than first zone), and cooperation zone (players not expected to receive ball within few second). 

\begin{figure}[ht]%
    \centering
    \subfloat[K-means]{{\includegraphics[ width=6cm]{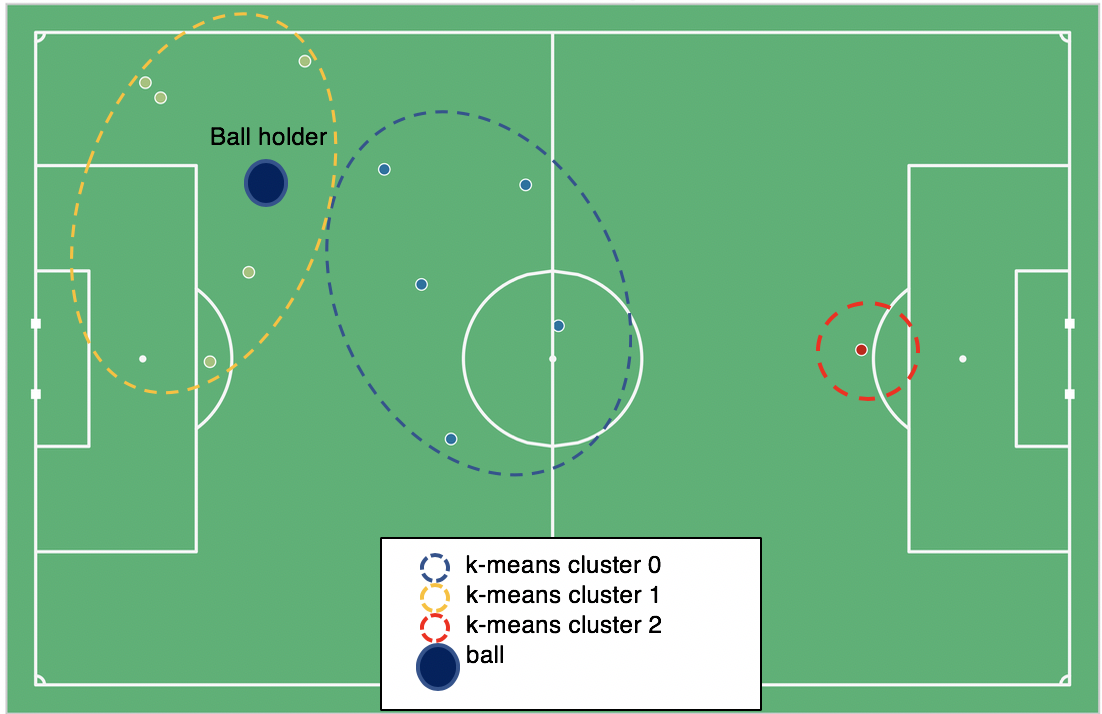}\label{k-means1} }}%
    \qquad
    \subfloat[Spectral clustering]{{\includegraphics[width=6cm]{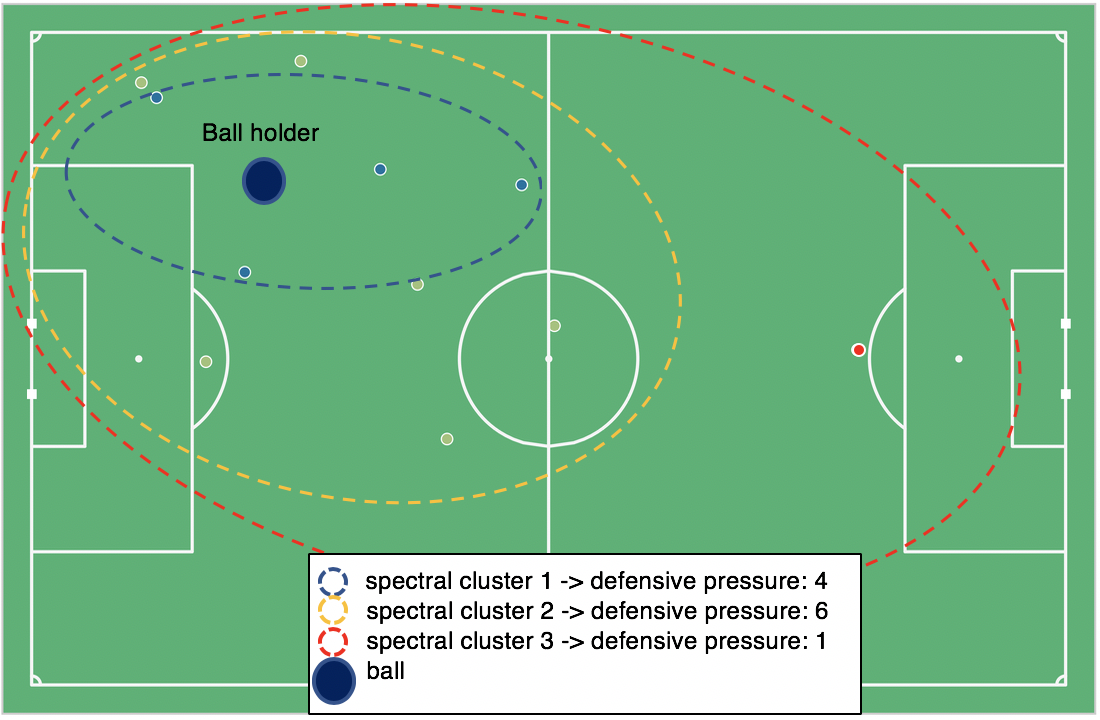}\label{spectral1} }}%
    \caption{Pressure model: number of opponent players in each zone/cluster is considered as pressure on ball holder in that zone.}%
    \label{clustering}%
\end{figure}

Our approach of modeling the pressure by the opposing team and considering game context in our valuation framework, matches the self-propelled particle model in grouping the opponents into several zones around the ball holder. To this end, we take advantage of a clustering method, keeping into consideration that opponents inside the clusters are not distributed spherically (according to their positions and velocities). K-means algorithm demonstrates a Pyrrhic victory, as it assumes that the clusters are roughly spherical and operates on Euclidean distance (Figure~\ref{k-means1}). But in soccer tracking data, such clusters are unevenly distributed in size. Thus, we experimented with spectral clustering to provide the number of opponents inside each cluster as an indicator of defensive pressure around the ball holder. We treated the positions $(x,y)$ and velocity $(v_x,v_y)$ of the opponents around the ball holder as graph vertices, and we constructed a k-nearest neighbors graph for each frame (5 neighbors in this work). In this graph, nodes are the opponent players' positions and velocities (direction and magnitude), and an edge is drawn from each position to its k nearest neighbors in the original space. The graph Laplacian is defined by the difference of adjacency and degree matrices. Then, we used K-means to perform clustering on vectors of the zero eigenvalues (connected components) from the Laplacian by setting the exact position (x,y) of the ball holder at each frame as the initial centroid of the clusters. Thus, each opponent player can be perfectly assigned to a spectral cluster (See Figure \ref{spectral1}).

Moreover, we experimentally selected the optimal number of clusters to be 3, using the elbow method by setting the metric to distortion (computes the sum of squared distances from each point to its assigned center) and inertia (sum of squared distances of samples to their closest cluster center). 
 Figure~\ref{clustering} depicts a frame of a specific match in our dataset with K-means on the top, and Spectral clustering on the bottom. Opponents from the away team are clustered into 3 groups. In the bottom Figure \ref{spectral1}, Cluster 1 (blue) includes 4 opponents who might immediately take the possession of the ball, cluster 2 (yellow) are opponents who might intercept the pass or dribble of ball holder, and cluster 3 (red) are opponents who cannot reach the ball in a few seconds. We compute the number of opponents for all frames of the matches in our dataset, and use them as the pressure feature throughout the rest of this work. The pseudo-code of the clustering algorithm for pressure measurement is provided in Algorithm~\ref{pseudo}.

\begin{algorithm}
	\caption{Defensive pressure measurement with Spectral clustering} 
	\label{pseudo}
	\begin{algorithmic}[1]
	    \State Set T: total number of frames, A: adjacency matrix, D: degree matrix, L: graph Laplacian. Initialize P: (opponent players data), Z: (connected components), C: (pressure clusters)
	    
		\For {$t\in T$}
		    \State $b\leftarrow (x_b, y_b, v_{xb}, v_{yb}) $ \Comment{Ball holder (b)}
			\For {$o=1,\ldots,11$}       
			    \State $p_o\leftarrow (x_o, y_o, v_{xo}, v_{yo}) $  \Comment{Each opponent player}
			    \State $p_{os}\leftarrow StandardScaler(p_o) $ 
			    \State $P.append(p_{os}) $ 
			 \EndFor
		    \State $A\leftarrow kneighbors\_graph(P)$	
		    \State $D\leftarrow diag(A)$
		    \State $L\leftarrow D-A$  \Comment{Graph Laplacian}
		    \State $vals , vecs \leftarrow eig(L)$
		    \State $eigenvecs\leftarrow vecs[:,sort(vals)]$ 
		    \State $eigenvals\leftarrow vals[sort(vals)]$
		    
		    \For {$i \in eigenvals$}       
		    \If {$abs(i) < 1e-5 $}   
		    \State $index \leftarrow argwhere(i)$
		    \State $z \leftarrow eigenvecs[:,index]$
		    \State $Z.append(z)$
		    \Comment{Connected components}
		    \EndIf
		   
			 \EndFor
			 
			 \State $C \leftarrow Kmeans(Z, init= b )$
			 \Comment{Cluster assignment}

		\EndFor
	\end{algorithmic} 
\end{algorithm}

\subsection{Selected features}

The InStat tracking data is one frame per second representation of positions for all the players including home and away.
As mentioned in 
Section~\ref{clustering-sec}, we have taken advantage of tracking data to calculate the velocities and opponents' location around the ball holder, and compute the defensive pressure in 3 different pressure zones to reduce dimensionality of the feature set. Another option was to avoid clustering the position features, and feed the network with the 44-dimensional ((x,y) for 22 players) normalized locations on the pitch. On the other hand, angle and distance to goal, time remaining, home/away, and body id can be directly calculated from event stream data. Table~\ref{features} shows the final list of our analyzed features used for our machine learning tasks in the following sections. Note that we either use location features (44-dimensional of exact players' locations) or pressure features (numbers of players in each cluster) to represent 3 state types in Table~\ref{cnn-table}. \textcolor{blue}{}

\begin{table}[h]
 \caption{Feature set}
 \centering
  \label{features}
  \begin{tabular}{|p{5cm}|p{3cm}|p{7cm}|}
  \hline

    \textbf{Feature set} & \textbf{Feature name} & \textbf{Description}  \\
     \hline
    hand-crafted & Angle to goal & the angle between the goal posts seen from the shot location\\
     \hline
     hand-crafted & Distance to goal & Euclidean distance from shot location to center of the goal line \\
     \hline
     hand-crafted & Time remaining & time remained from action occurrence to the end of match half \\
     \hline
     hand-crafted & Home/Away & action is performed by home or away team?  \\
    \hline
     hand-crafted & Action result & successful or unsuccessful  \\
    \hline
    hand-crafted  & Body ID & action is performed by head? body? foot?  \\
    \hline
    contextual: clustered locations & Pressure in zone 1 & number of opponents in first cluster  \\
    \hline
    contextual: clustered locations & Pressure in zone 2 & number of opponents in second cluster  \\
    \hline
    contextual: clustered locations & Pressure in zone 3 & number of opponents in third cluster  \\
    \hline
    contextual: exact locations & locations & 44-dimensional exact locations (x,y) of opponents  \\
    \hline
       
\hline
  \end{tabular}
\end{table}

\subsection{Possession input representation}

State representation is one of the most challenging steps in soccer analytics due to the high-dimensional nature of the datasets. We describe each game state by generating the most relevant features and labels to them. To this end, we define a different set of features, i.e., hand-crafted and contextual (Table~\ref{features}), and 3 types of state representation (Table~\ref{cnn-table}).

For each of the state types (I, II, III), we demonstrate the state as the combinations of different features vector $X$ (see Tables~\ref{features},\ref{cnn-table}), and one-hot representation of the action $A$ for all the actions inside each possession, excluding the ending action. Thus, the varying possession length is the number of actions inside a possession, excluding the ending one. Then, the state is a 2 dimensional array, with the first dimension of possession length: (varying for each possession), and second dimension of features number. Therefore, a $m^{th}$ state/possession with length of $n$ can be represented as $S_m = [[X_0,A_0], [X_1,A_1],...,[X_{n-1},A_{n-1}]]$.

Due to the complex and spatiotemporal nature of the dataset, we select the best representation of the state through an experimental process. To do this, we train the spatiotemporal models on three different state types. State type (I) ignores the players' locations and only reflects the occurred actions in addition to the hand-crafted features of each action. State type (II) is a high-dimensional representation that considers exact players' locations besides the actions and their corresponding hand-crafted features. In the state type (III), we handled the curse of dimensionality of type (II) by clustering the locations as shown in Section~\ref{clustering-sec}. See Table~\ref{cnn-table} for more details on states.

\subsection{CNN-LSTM architecture for deriving behavioral policy}
\label{cnn-lstm}
\begin{figure}[h]
  \includegraphics[ width=\textwidth]{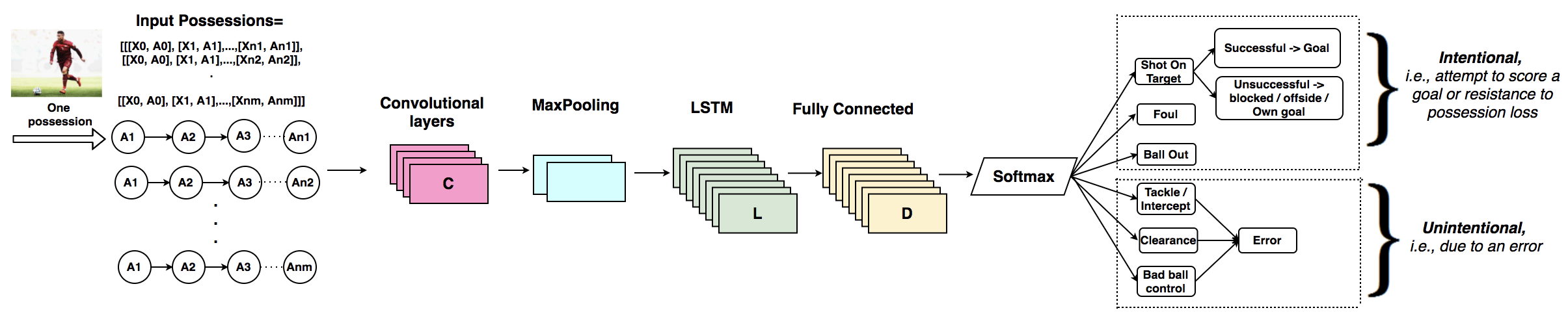}
  \caption{CNN-LSTM network structure for classification of the possessions, i.e., action sequences. Input possessions represent both state features vector $(X)$, and one-hot vector of actions $(A)$, excluding the ending action. There are $m$ possessions in the dataset, with varying lengths of $\{n_1,n_2,...,n_m.\}$. The output is the predicted class (ending action) of the possession, along with the estimated probabilities of alternative ending actions. }
  \label{cnn}
\end{figure}

In the soccer event dataset, each possession is represented by a sequence of actions. We aim to classify these possessions (and show the result only for the home team) based on their ending actions. Thus, each possession should be terminated by the following classes: 
1) Shot (goal or unsuccessful), 2) Ball out, 3) Foul, 4) Errors (possession loss due to inaccurate pass, bad ball control, or tackle and interception by opponent). Note that foul and ball out actions are only performed by the home team. Thus, if the possession is terminated by any action from the opponent, including foul and ball out, we classify them as error.
To this end, we utilize the classification capability of sequence prediction methods. 
In order to handle the spatiotempral nature of our dataset, we needed a sophisticated model and best feature set, which could optimize the prediction performance. Thus, model selection was the core task of this study. We first created appropriate state dimensions suitable for each model by reshaping the state inputs, then fed our reshaped arrays with the 3 state types to the following networks: 3D-CNN, LSTM, Autoencoder-LSTM, and CNN-LSTM, to compare their classification performance (See Table~\ref{cnn-table}). Validation split of 30\% of consecutive possessions is used to evaluate during training, and cross entropy loss on train and validation datasets is used to evaluate the model. As the table suggests, CNN-LSTM \cite{jeff2016} trained on state type (III) outperforms other models in terms of accuracy and loss. Thus, the necessity of the exact location of the players can be rejected and sufficiency of pressure features can be proved in this analysis. Although the Autoencoder-LSTM accuracy trained on state types (II and III) is quite similar to CNN-LSTM, its relatively large inference time and trainable parameters make the implementation more strenuous and expensive.
Thus, we continued the rest of the analysis by developing a CNN-LSTM network \cite{jeff2016}, using CNN for spatial feature extraction of input possessions, and LSTM layer with 100 memory units (smart neurons) to support sequence prediction and interpret features across time steps. Figure~\ref{cnn} depicts the architecture of our network. Since our input possessions (possession array) have a three dimensional spatial structure, i.e., first dimension: number of possessions, second dimension: dynamic possession length (maximum=10), third dimension: number of features, CNN is capable of picking invariant features for each class of possession. Then, these learned consolidated spatial features are fed to the LSTM layer. Finally, the dense output layers are used with softmax activation function to perform our multi-classification task. 

Note that the feature vector $X$ has a fixed length for each individual action, but varying for all actions in the state (because possession length or number of actions varies). This is one of the main challenges in our work as most machine learning methods require fixed-length feature vectors. 
In order to address the challenge of dynamic length of possession features (second dimension of possession input array), we use truncating and padding. We mapped each action in a possession to an 11 length real-valued vector. Also, we limit the total number of actions in a possession to 10, truncating long possessions and we pad the short possessions with zero values. In this case, we will have a fixed length of sequences through the whole dataset for modeling.

Consequently, this network estimates the categorized probability distribution over actions for any given possession, parameterized by $\theta$. Through the rest of the paper, we denote this probability distribution as $q(x;\theta)$.

\begin{table}[h]
\centering
 \caption{Classification performance of different design choices for spatiotemporal analysis. (Inference times are the average running times over 20 iterations of training using a server enriched with Tesla K80 GPU)}
    \begin{adjustbox}{width=1.\textwidth,center=\textwidth}
  \label{cnn-table}
        \begin{tabular}{|>{\centering\arraybackslash}m{6.2cm}|>{\centering\arraybackslash}m{2.55cm}|>{\centering\arraybackslash}m{2cm}|>{\centering\arraybackslash}m{3cm}|>{\centering\arraybackslash}m{1.5cm}|>{\centering\arraybackslash}m{1.5cm}|>{\centering\arraybackslash}m{1.5cm}|>{\centering\arraybackslash}m{1.5cm}|}
  \hline

  \textbf{State representation} & \textbf{State type} & \textbf{Initial state dimension} & \textbf{Spatio temporal model} & \textbf{Accuracy} & \textbf{Loss}&\textbf{Inference time} & \textbf{Parameters}\\
  \hline
  
  \multirow{ 4}{*}{hand-crafted(6) + actions(11)} & \multirow{ 4}{*}{(I).non-contextual} & \multirow{ 4}{*}{[28054,10,17]} & 3D-CNN & 61\% & 0.73 & 0.15s & 52,321 \\
   & && LSTM & 65\% & 0.72 & 0.09s & 42,001 \\
   && & Autoencoder-LSTM & 69\% & 0.71 & 3.7s & 89,099 \\
    & && CNN-LSTM & 68\% & 0.71 & 0.25s & 32,456 \\
    
  \hline
  
  \multirow{ 4}{*}{hand-crafted(6) + locations(44) + actions(11)}  & \multirow{2}{*}{(II).contextual,}
   & \multirow{ 4}{*}{[28054,10,61]} & 3D-CNN & 72\% & 0.63 & 1.71s & 172,981 \\
   && & LSTM & 68\% & 0.69 & 0.81s & 151,034 \\
   &\multirow{2}{*}{high-dimensions} & & Autoencoder-LSTM & 80\% & 0.59 & 12.32s & 350,024 \\
    && & CNN-LSTM & 75\% & 0.65 & 1.22s & 152,211\\
    
  \hline
  
  \multirow{ 4}{*}{hand-crafted(6) + pressures(3) + actions(11)} & \multirow{2}{*}{(III).contextual,} & \multirow{ 4}{*}{[28054,10,20]} & 3D-CNN & 73\% & 0.63 & 0.31s & 61,211 \\
   && & LSTM & 71\% & 0.63 & 0.11s & 50,804 \\
   &\multirow{2}{*}{reduced-dimensions}& & Autoencoder-LSTM & 79\% & 0.59 & 5.02s & 92,022 \\
    && & \textbf{CNN-LSTM} & \textbf{81}\% & \textbf{0.56} & 0.51s & 56,036 \\
    
    \hline

\hline
        \end{tabular}
    \end{adjustbox}
\end{table}

\section{Off-policy reinforcement learning}
\label{rl-sec}
Most RL methods require active data collection, where the agent actively interacts with the environment to get rewards. Obviously, this situation is impossible in our soccer analytics problem, since we are not able to modify the players' actions. Thus, our study falls right into the category of batch RL. In this case, we will not face the exploration vs. exploitation trade-off, since the actions and rewards are not sampled randomly, but they are sampled from the real world (players' actions in a match) before the learning process. Moreover, our network learns a better target policy from a fixed set of interactions.


Before the learning process, the players selected some ending actions according to some non-optimal (behavioral) policy. We aim to use those selected actions and acquired rewards to learn a better policy. 
Therefore, we prepared our dataset of transitions in the form of $<$current observation, action, reward, next state$>$ for learning a new policy. 
Through the end of the paper, we use the notation and definition shown in Table~\ref{notations}.

\begin{table}[h]
 \caption{Notations}
 \centering
  \label{notations}
  \begin{tabular}{|p{5cm}|p{10cm}|}
  \hline
    \textbf{Notation} & \textbf{Definition}  \\
     \hline
    State: ($s$) & Sequence of actions and their features in a possession of each team, excluding the ending action \\
     \hline
    Action: ($a$) & Ending action of each possession, which leads to state transition \\
     \hline
    Episode: ($\tau$) & Sequence of possessions of the home team, until they lose the possession, or end it with a shot, denoted by $\tau = \{s(\tau , t), a(\tau , t)\}_{t=1}^T$   \\
     \hline
    Reward: $r(s_t,a_t)$ & Reward acquired from each ending action at the end of a possession   \\
    \hline
    Episode reward: $r(\tau)$  & Sum of rewards (expected goals) for each episode: $r(\tau)= \sum_{t=1}^Tr(s_t,a_t)$  \\
    \hline
    Return: $R$  & Cumulative discounted and normalized reward  \\
    \hline
    Target policy distribution: $p(x)$ & Learned policy (probability distribution of actions from the policy network)   \\
    \hline
    Behavior policy distribution: $q(x)$ & Actual policy (probability distribution of actions collected off-line from a real match)   \\
    \hline
    $n_i$ & Length (number of actions) in $i^{th}$ possession \\
    \hline
    $m$ & Total number of possessions  \\
    \hline
       
\hline
  \end{tabular}
\end{table}

\subsection{Action reward function}

In this section, we aim to estimate the reward acquired for the ending actions. Owing to the complex and sparse environment of soccer games, it is tedious to design the perfect reward function. In general though, every team desires to be in the most precious states, i.e., with maximum probability of goal scoring, as much as possible. 

In the soccer dataset (for either the home or the away team), each episode starts from the moment that the team acquires the possession of the ball, and it terminates when the team either loses the possession (loss), or it ends up shooting (win). 
According to the Markovian possession model in Section~\ref{sec:markov}, we have the set of ending actions (out, foul, shot, error) which are leading to state transitions. We estimated the probabilities of a possession belonging to the shot class in Section~\ref{cnn-lstm} with the help of a CNN-LSTM network. In order to define the value of each possession, we need to define the following concepts:

\begin{itemize}
\item \textbf{$P(shot | X)$:} computed by CNN-LSTM, is the probability of possession belonging to the shot class, given the features of the possession.
\item  \textbf{$P(goal | shot , X)$: }is the probability of goal scoring, assuming that a possession belongs to the shot class, and given the shot features. This is the same concept as the state-of-the-art expected goal (xG) model that classifies shots to goal and no-goal. In this work, we have computed xG using logistic regression and show its higher performance with 5-fold cross-validation in comparison with other classifiers in Table~\ref{classifier}. (Details in Appendix~\ref{xg})
\end{itemize}

It has become evident that higher $P(shot | X)$ indicates a higher chance of a shot. Accordingly, the higher $P(goal | shot , X)$ shows a better chance of goal scoring. Thus, the multiplication of these two terms will give us the Possession Value (PV) in state s, denoted in \eqref{eq1}. The Bayesian formula for this equation is provided in Appendix~\ref{bayes}.

\begin{equation}
\label{eq1}
PV(s) = P(shot | X) P(goal | shot , X)
\end{equation}

Now we define the rewards acquired by each ending action in a possession. The most precious actions in critical situations have 2 criteria: 1) prevent possession loss, 2) save the possession for the team, and lead transition to a more valuable possession with higher PV. Thus, we present our reward function as depicted in \eqref{eq2}:

\begin{equation}
\label{eq2}
    \begin{split}
r(s,a) = \left\{
    \begin{array}{lll}
        PV(s) & \mbox{if a is a shot;}\\
        PV(s^\prime)-PV(s) & \mbox{else and } s,s^\prime \in \mbox{same team;}\\
        -0.1 & \mbox{else,}
    \end{array}
\right.
    \end{split}
\end{equation}
where $r(s,a)$ is the reward when the state changes from $s$ to $s^\prime$ by taking action $a$.
Our proposed reward function computes the immediate reward by the arbitrary action that each player performed. Choosing the shot, the player receives the PV of the possession. If he performs any action other than shot (e.g., ball out or foul), but the next possession is still for his team, the model computes the PV of the next possession and compares it to the current possession. On the other side, if he performs any action leading to possession loss (e.g., bad ball control, inaccurate pass, tackle and interception by opponent), he should receive a negative reward. In this work, -0.1 proved to be the best reward of possession loss to confirm the convergence of the policy network. Moreover, the sum of $r(s,a)$ at each time-step throughout the whole episode is the indicator of expected goal for the team. Thus, the control objective is to maximize the expected goal of the teams.

\begin{table}[h]
 \centering
 \caption{Expected Goal computation performance}
  \label{classifier}
  \begin{tabular}{p{4.5cm}|p{1cm}|p{1cm}}
  \hline
    \textbf{Classifier} & \textbf{Brier} & \textbf{AUC}  \\
     \hline
    XGBoost & 0.014 & 0.765\\
    Random Forest & 0.014 & 0.759\\
    SVM & 0.015 & 0.733\\
    \textbf{Logistic Regression} & \textbf{0.012} & \textbf{0.798}\\
       
\hline
  \end{tabular}
\end{table}   

\subsection{Training protocol and return}

For each state, the network needs to decide about performing the appropriate action with the corresponding parameter gradient. The parameter gradient tells us how the network should modify the parameters if we want to encourage that decision in that possession in the future.
We modulate the loss for each action taken at the end of a possession according to their eventual outcome, since we aim to increase the log probability of successful actions (with higher rewards) and decrease it for the unsuccessful actions.

We define discounted reward (return) for episode $\tau$ in \eqref{eq3}.

\begin{equation}
\label{eq3}
R(\tau) = \frac{\sum_{t=0}^{\infty} \gamma^{t}\times r(s_{\tau,t}, a_{\tau,t})}{\sum_{t=0}^{\infty} \gamma^{t}}
\end{equation}

where $\gamma$ is a discount factor (Appendix~\ref{gamma}), and $r$ is the estimated rewards (expected goals) for time-step $t$ after standardization to control the gradient estimator variance. $R$ shows that the strength of encouraging a sample action at the end of a possession is the weighted sum of rewards (expected goals) afterwards. In this work, we constrain the look ahead to the end of the episodes. 

\subsection{Policy gradient}
Policy gradient (PG) is a type of score function gradient estimator. Using PG, we aim to train a policy network that directly learns the optimal policy by learning a function that outputs the best action to be taken in each possession. 


The CNN-LSTM network in Section~\ref{cnn-lstm} estimated the behavioral probability distribution over actions (shot, out, foul, error) for any given possession denoted by $q(x)$. This categorized probability distribution demonstrates some nonrandom, regular, and non-optimized policies obeyed by the players and possibly dictated by coaches through the matches. 
In order to find a better policy, which optimizes the expected goal of episodes, we need to train the network. We call this network a target policy network $p(x)$. The training is done with the help of gradient vector, which encourages the network to slightly increase the likelihood of highly positive rewarding actions, and decrease the likelihood of negative ones. 
We seek to learn how the distributions should be shifted (through its parameter $\theta$), in order to increase the reward of the taken actions. 

In the general case, we have the expression of form: \[E_{x\sim p(x;\theta)}[f(x)],\] in which $f(x)$ is our return, and $p(x;\theta)$ is our learned policy. In our soccer problem, this expression is an indicator of expected goals in each episode through the whole match. In order to maximize the expected goals, we need to compute the gradient vector $\nabla_\theta \log p(x;\theta)$ as follows:

\begin{gather*} 
\nabla_\theta E_x[f(x)]=\nabla_\theta \sum_x p(x)f(x) \Leftarrow \mbox{expectation of return} \\
=\sum_x \nabla_\theta p(x)f(x) \Leftarrow \mbox{swap sum and gradient} \\
=\sum_x p(x) \frac{\nabla_\theta p(x)}{p(x)}f(x) \Leftarrow \mbox{multiplying and dividing by p(x)} \\
=\sum_x p(x) \nabla_\theta \log p(x)f(x) \Leftarrow \mbox{because } \nabla_\theta \log (z) = \frac{1}{z}\nabla_\theta z \\
=E_x[f(x)\nabla_\theta \log p(x)] \Leftarrow \mbox{expectation }
\end{gather*}

But the PG is considered to be on-policy, i.e., training samples are collected according to the target policy. This situation is not valid in our off-line setting and we encounter out-of-distribution actions. Thus, we need to reformulate the PG as in \eqref{eq4} considering importance weight $\frac{p(x)}{q(x)}$ (proof in Appendix~\ref{pg-off}).

\begin{equation} 
\label{eq4}
\nabla_\theta E_x[f(x)] = E_x[\frac{p(x)}{q(x)}f(x)\nabla_\theta \log p(x)]
\end{equation}

Gradient vector $\nabla_\theta \log p(x;\theta)$, is the gradient that computes a direction in the parameter space leading to an increase of the probability assigned to $x$. Consequently, high rewarding actions will tug on the probability density stronger than low rewarding actions. Therefore, by training the network, the probability density would shift around in the direction of high rewarding actions, making them more likely to occur.

\subsection{Off-policy training}

Our soccer analysis problem in this work falls right into the category of the off-policy variant of RL methods. In this method, the agent learns (trains and evaluates) solely from historical data, without online interaction with the environment. 



\begin{figure*}
  \centering
  \includegraphics[width=15cm]{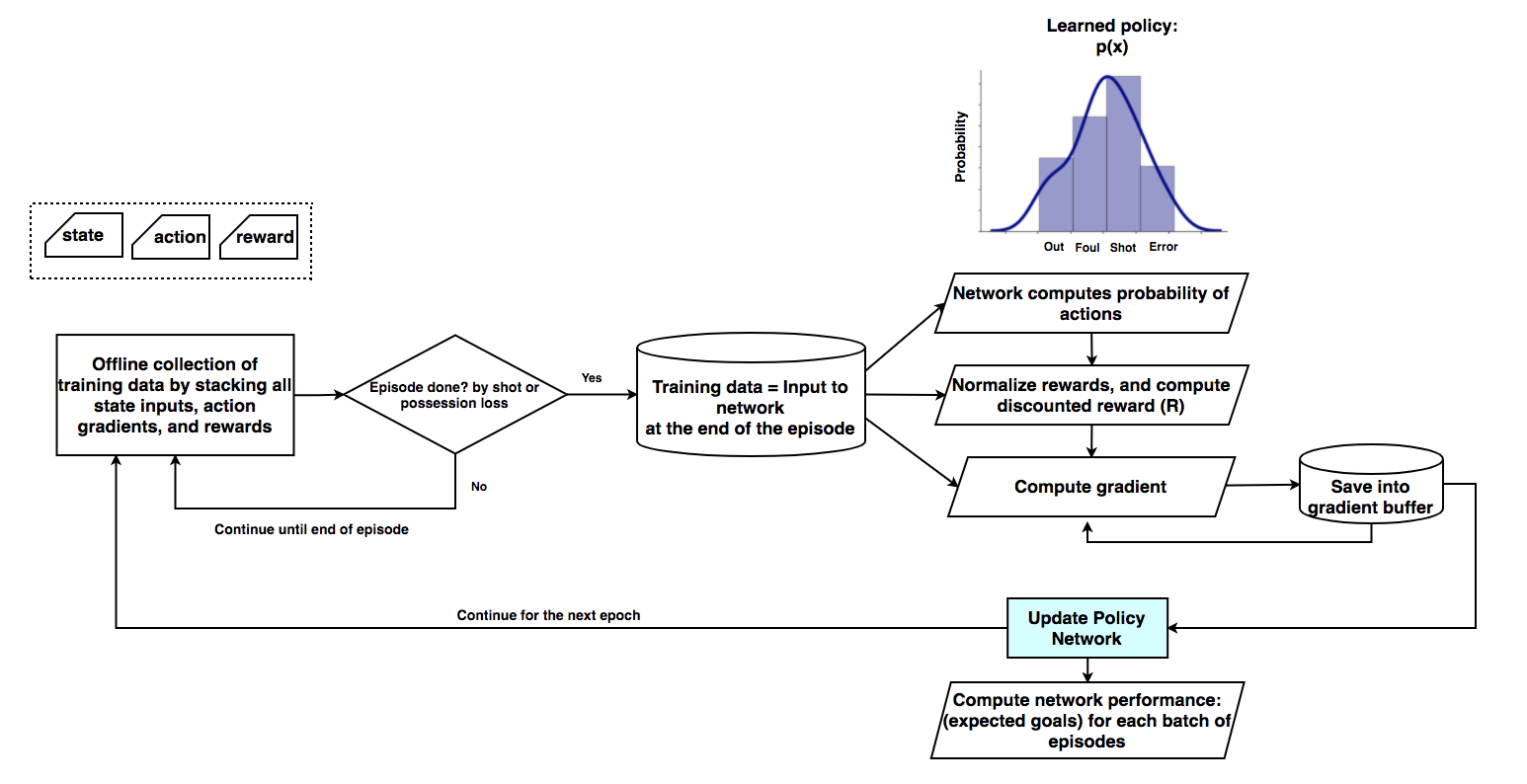}
  \caption{Offline training workflow of policy network}
  \label{pg}
\end{figure*}

Figure~\ref{pg} illustrates our training workflow of the policy network, with off-line data collection, and gradient computation.

\section{Experimental results}
\label{ope-sec}
It is a challenging task to evaluate our implemented framework, as there is no ground truth method for action valuation or optimizing the policy in soccer. Therefore, we evaluate the performance of our proposed framework with an eye towards two questions: 1) How well our trained network can maximize the expected goal in comparison to the behavioral policy? We answer this question by the off-policy policy evaluation (OPE) method. 2) What is the intuition behind the selected actions of our target policy? We elaborate on this by providing three scenarios of the most critical situations in a particular match from the dataset. The structure of the datasets used in this study is provided in Appendix~\ref{data}.
\subsection{Off-policy policy evaluation with importance sampling and doubly robust methods}

Applying the off-policy method in our soccer analysis problem, we faced the following challenge: while training can be performed without a real robot (simulator), the evaluation cannot, because we cannot deploy the learned policy in a real soccer match to test its performance. This challenge motivated us to use off-policy policy evaluation (OPE), which is a common technique for testing the performance of a new policy, when the environment is not available or it is expensive to use.

With OPE, we aim to estimate the value and performance of our newly optimized policy based on the historical match data collected by a different behavioral policy obeyed by the players.
For this aim, we use the importance sampling method used by different works such as Teng et al. \cite{Teng2019} and doubly robust in \cite{DBLP:journals/corr/abs-1802-03493} and \cite{DBLP:journals/corr/JiangL15}. They take samples from behavioral policy $q(x)$ to evaluate the performance of target policy $p(x)$. The workflow of the evaluation with importance sampling is sketched in Figure~\ref{sampling} of Appendix~\ref{imp}, and details of the doubly robust are provided in Appendix~\ref{dr}. Moreover, the input dataset format to the OPE is shown in Table~\ref{policy} of Appendix~\ref{data-qppendix}.



\subsection{Experiments}
We used the 104 games on 3 state types to train our policy, and evaluate it by the OPE methods. In this section, we demonstrate the performance of the obtained policy and compare it to the behavior policy on different state representations. Then we mention three scenarios and analyze the performance of our policy versus the real players' actions.

\begin{figure}[h]
\centering
  \includegraphics[ width=.8\textwidth]{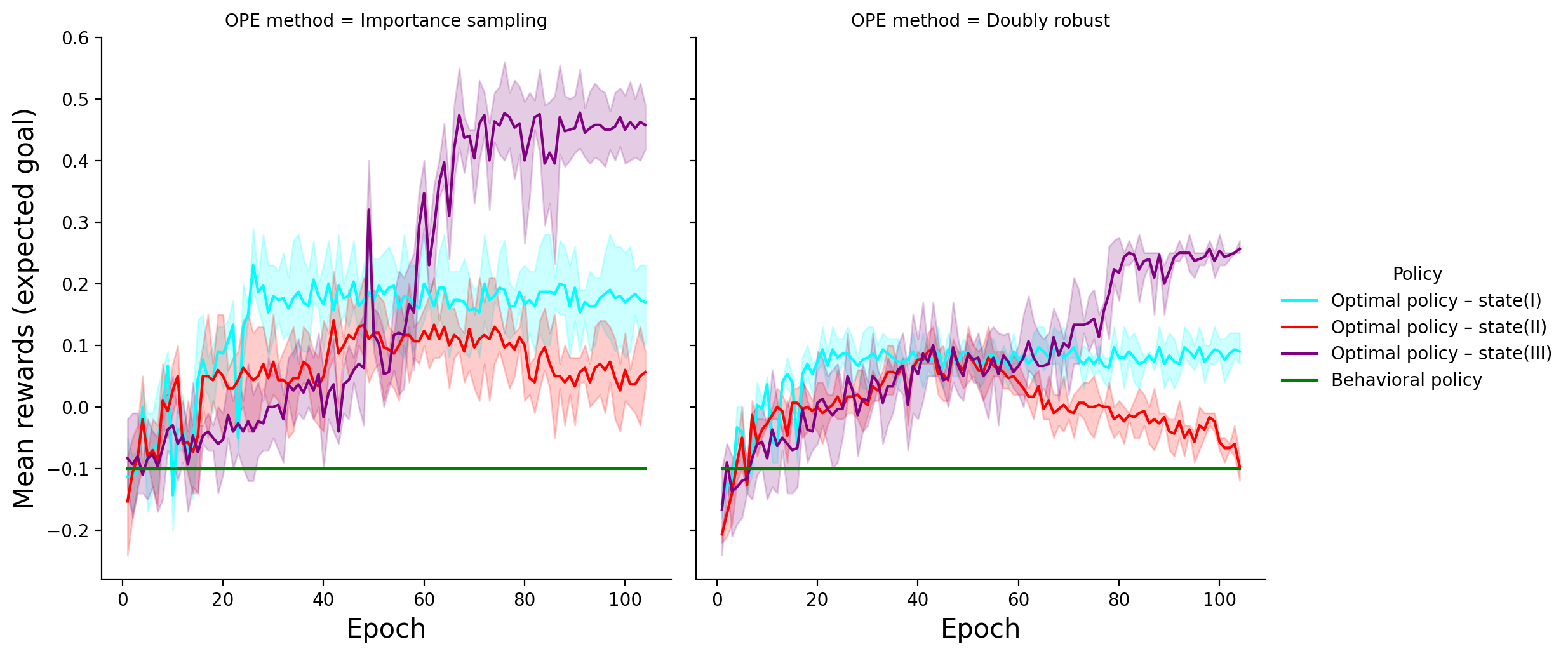}
  \caption{Off-policy policy evaluation on the 3 state representations of the trained models with importance sampling and doubly robust methods, and compare it to behavioral policy. Shaded region represents standard deviation over 104 game rollouts}
  \label{models}
\end{figure}

Figure \ref{models} shows mean rewards over 100 epochs of the trained policy network using the different proposed state representations (see Table~\ref{cnn-table}) evaluated by importance sampling and doubly robust methods. As the Figure reveals, both OPE methods show that our proposed state representation type(III) (purple line) could let the policy network converge after sufficient epochs. Particularly under state(III), the policy network converges after about 70 epochs evaluated by importance sampling, and around 80 epochs evaluated by doubly robust. On the other hand, mean rewards curves under state(I) are quickly converging (due to their low-dimensional input) to a relatively lower mean rewards, and mean rewards curves under state(II) are failing to converge (due to their high-dimensional and complex input structure). Thus, the results obviously prove that our proposed state representation (III) is outperforming than other types. 
Using importance sampling as the better evaluator of the optimal policy with state (III), any model after epoch number 70 is suitable for going into deployment by the football club for analysis. We can see that the acquired reward (expected goal) by the trained policy is around 0.45 with some variances on average of all 104 games. This figure also shows that the optimized policy (purple line) is outperforming the mean rewards by behavioral policy (green line), which is about -0.1 for all the matches. 


\begin{figure}[h]
  \centering
  \includegraphics[ width=5cm]{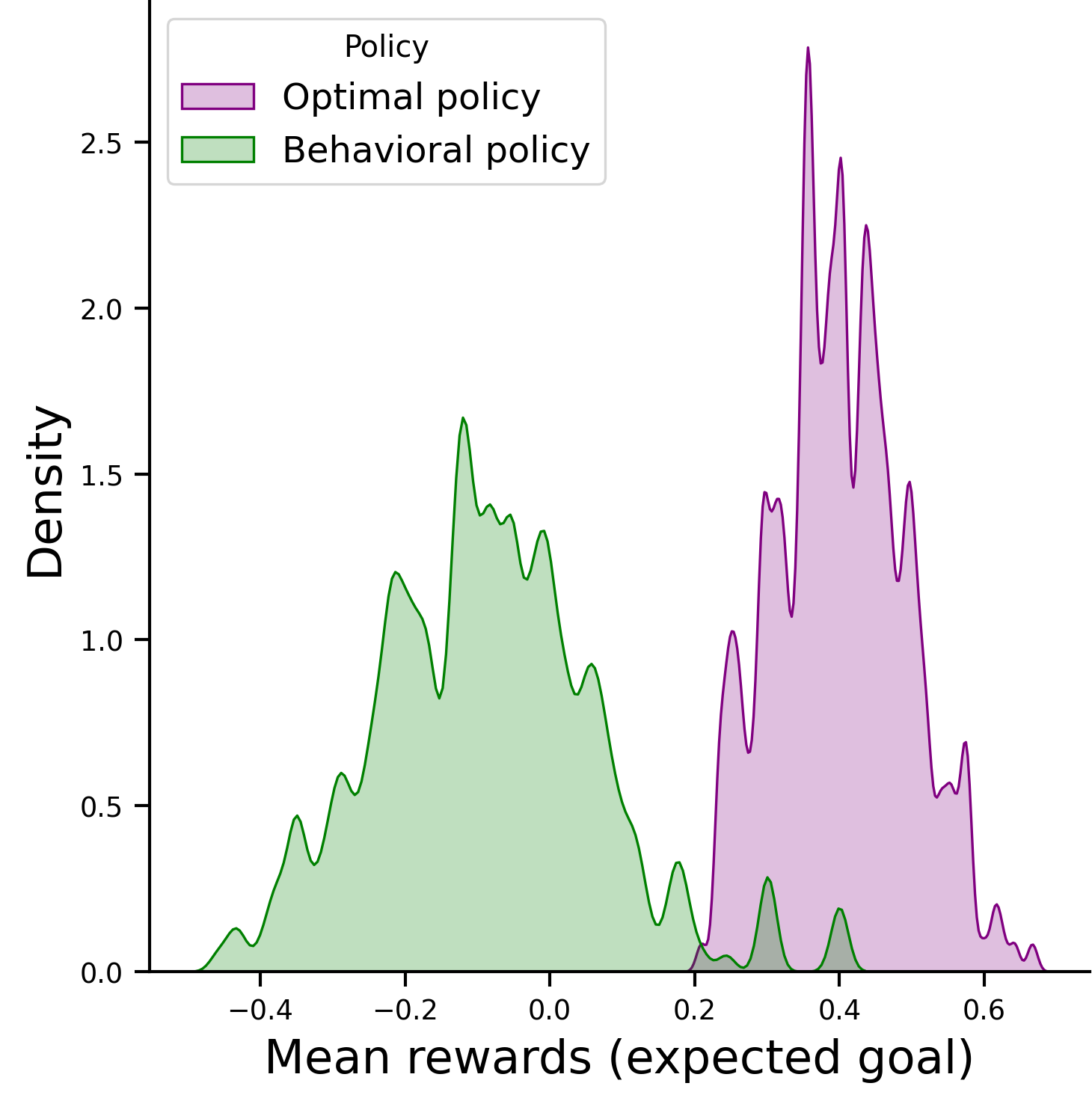}
  \caption{KDE of mean rewards for all episodes of 104 games, acquired by behavioral and optimal policy (evaluated by importance sampling)}
  \label{rewards}
\end{figure}

Moreover, Figure~\ref{rewards} compares the Kernel density estimation (KDE) of the mean rewards by behavioral and optimal policies for all matches, evaluated by OPE. As it is shown, the density of the optimized policy has moved to the positive side and clearly has improved over the behavior policy. It also has a smaller variance compared to the behavior policy.

\begin{figure}[h]
    \centering
    
    \subfloat[Scenario1: (performed action: foul), (optimal action: shot)]{{\includegraphics[height=4cm, width=5cm]{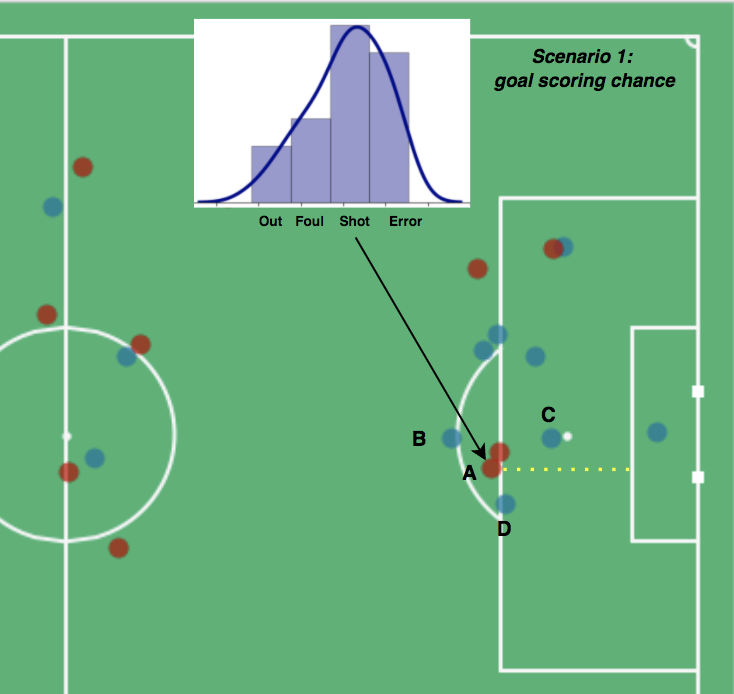}\label{scenario1}}}%
    \qquad
    \subfloat[Scenario2: (performed action: error/tackle by opponent), (optimal action: foul)]{{\includegraphics[height=4cm, width=5cm]{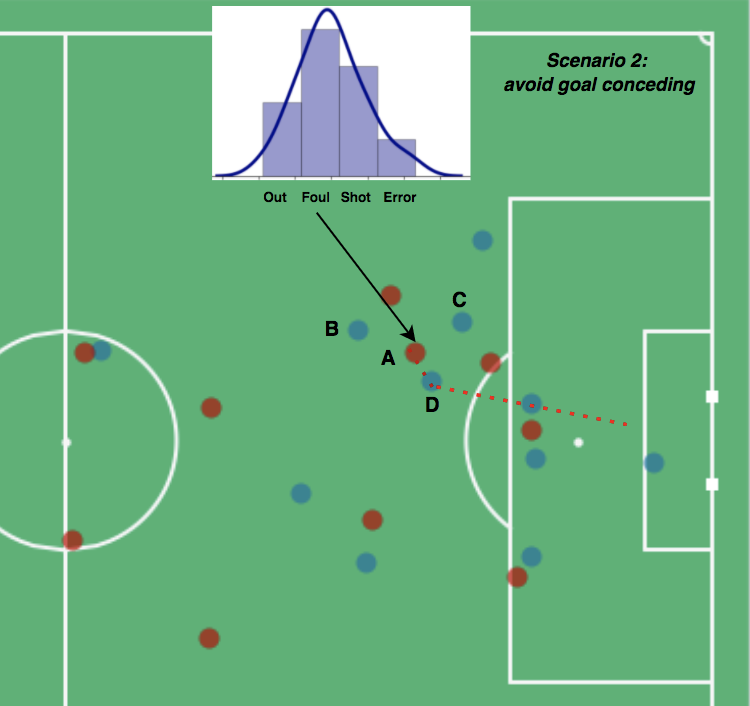}\label{scenario2} }}%
    \qquad
    \subfloat[Scenario3: (performed action: error/bad ball control), (optimal action: out)]{{\includegraphics[height=4cm, width=6cm]{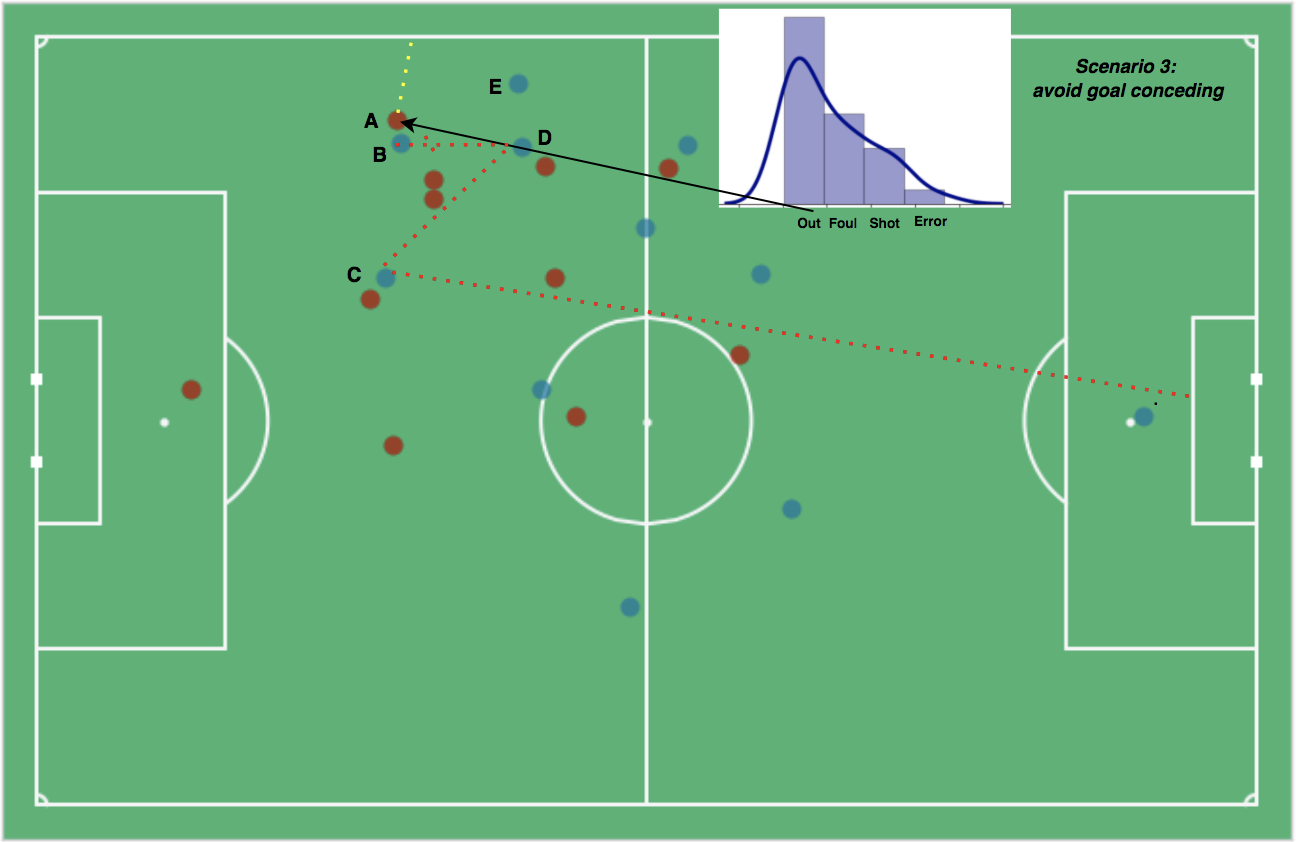}\label{scenario3} }}%
    \caption{Three scenarios of critical situations in a match. Red dots are home team players, and blues are away. Black arrow shows the ball holder. Yellow dashed lines show the optimal trajectory of ball, if the player was following optimal policy. Red dashed lines show the actual and non-optimal trajectory of ball by the performed action of player in the match. The probability distribution shows the optimal output of our trained policy network. }%
    
    \label{scenarios}%
\end{figure}

Now, we consider some scenarios to see how the optimized policy works compared to the behavior policy.
Figure \ref{scenarios} sketches 3 different scenarios in the critical situations of a particular match in our dataset, when there is no chance of pass or dribble for the ball holder. Thus, the ball holder needs to decide about the 3 intentional options (shot, out, foul), or submit the ball to the opponent by an error. The scenarios of the performed action by the player, and the proposed action by policy network are the following.

\textbf{Scenario 1: home player missed goal scoring opportunity: }
Figure~\ref{scenario1} shows the $24^{th}$ episode of the match. Player A from the home team stops a long sequence of passes by committing a foul and he gets the reward of -0.16. In this second, there is high pressure from away players (B,C,D). So A tries a foul to prevent possession loss. Then player D from away team gets the ball, but immediately loses it due to an inaccurate pass. As claimed before, the unsuccessful touches of opponent in less than 3 consecutive actions are not considered as possession loss. Thus, the possession is kept for the home team after committing a foul by A. Although the possession was kept for the home team after this action, the policy network suggests shooting the ball instead of committing the foul. So player A could gain the reward (expected goal) of 0.4, meaning that the probability of goal scoring was 0.4, and he missed this opportunity.

\textbf{Scenario 2: goal conceding:}
Figure~\ref{scenario2} shows the $73^{th}$ episode of the match. Player A from home loses the possession by error (tackle by D) and he gets -0.1 reward. The next possession belongs to the away team, and they score a goal (red trajectory in the figure). The policy network assigns a higher probability for foul in this situation instead of this inaccurate pass (error), so there was a chance of saving the possession for A, and avoid goal conceding for the home team. 

\textbf{Scenario 3: goal conceding: }
Figure~\ref{scenario3} shows the $90^{th}$ episode of the match. Player A from home loses the possession due to bad ball control and high pressure from B,D,E, and gets -0.1 of reward. The next possession belongs to away players and they score a goal (red trajectory). The policy network surprisingly suggests sending the ball out in this situation, so those home players could probably save the possession and avoid goal conceding. 

\section{Conclusion}
\label{conclusion}

We proposed a data-driven deep reinforcement learning framework to optimize the impact of actions, in the critical situations of a soccer match. In these situations, the player cannot pass the ball to a teammate, or continue with dribbling. Thus, the player can only commit a foul, send the ball out, shoot it, or if not skilled enough, she/he would lose the ball by a defensive action of the opponent. Our framework built on a training policy network will help the players and coaches to compare their behavioral policy with the optimal policy. More specifically, sports professionals can feed any state with the proposed possession features and state representation to find the optimal actions. We conducted experiments on 104 matches and showed that the optimal policy network can increase the mean rewards to 0.45, outperforming the gained expected goals by the behavioral policy, which is -0.1. To the best of our knowledge, this work constitutes the first usage of off-policy policy gradient reinforcement learning to maximize the expected goal in soccer games. A direction for future work is to expand the framework to evaluate all on the ball actions of the players, including passes and dribbles.   

\section*{Acknowledgment}
Project no. 128233 has been implemented with the support provided by the Ministry of Innovation and Technology of Hungary from the National Research, Development and Innovation Fund, financed under the FK\_18 funding scheme. The authors thank xfb Analytics\footnote{http://www.xfbanalytics.hu/} for supplying event stream and tracking data used in this work.

\bibliographystyle{IEEEtran}
\bibliography{template}

\begin{thebibliography}{10}
\providecommand{\url}[1]{#1}
\csname url@samestyle\endcsname
\providecommand{\newblock}{\relax}
\providecommand{\bibinfo}[2]{#2}
\providecommand{\BIBentrySTDinterwordspacing}{\spaceskip=0pt\relax}
\providecommand{\BIBentryALTinterwordstretchfactor}{4}
\providecommand{\BIBentryALTinterwordspacing}{\spaceskip=\fontdimen2\font plus
\BIBentryALTinterwordstretchfactor\fontdimen3\font minus
  \fontdimen4\font\relax}
\providecommand{\BIBforeignlanguage}[2]{{%
\expandafter\ifx\csname l@#1\endcsname\relax
\typeout{** WARNING: IEEEtran.bst: No hyphenation pattern has been}%
\typeout{** loaded for the language `#1'. Using the pattern for}%
\typeout{** the default language instead.}%
\else
\language=\csname l@#1\endcsname
\fi
#2}}
\providecommand{\BIBdecl}{\relax}
\BIBdecl

\bibitem{instat}
``Instat,'' \url{https://instatsport.com/}.

\bibitem{wyscout}
``Wyscout,'' \url{https://wyscout.com/}.

\bibitem{statsbombs}
``Statsbomb,'' \url{https://statsbomb.com/}.

\bibitem{stats}
``Stats,'' \url{https://www.statsperform.com/}.

\bibitem{opta}
``Opta,'' \url{https://www.optasports.com/}.

\bibitem{Fernandez2019}
J.~Fernandez, L.~Bornn, and D.~Cervone, ``Decomposing the immeasurable sport: A
  deep learning expected possession value framework for soccer,'' in \emph{MIT
  Sloan Sports Analytics Conference}, 2019.

\bibitem{Fernandez2020}
F.~P. Alguacil, J.~Fernandez, P.~P. Arce, and D.~Sumpter, ``Seeing in to the
  future: using self-propelled particle models to aid player decision-making in
  soccer,'' in \emph{MIT Sloan Sports Analytics Conference}, 2020.

\bibitem{fernandez2021}
J.~Fernandez and L.~Born, ``Soccermap: A deep learning architecture for
  visually-interpretable analysis in soccer,'' in \emph{ECML PKDD}, 2020.

\bibitem{gyarmati2016qpass}
L.~Gyarmati and R.~Stanojevic, ``Qpass: a merit-based evaluation of soccer
  passes,'' in \emph{KDD Workshop on Large-Scale Sports Analytics}, 2016.

\bibitem{Tom2019}
T.~Decroos, L.~Bransen, J.~V. Haaren, and J.~Davis, ``Actions speak louder than
  goals: Valuing player actions in soccer.'' in \emph{ACM KDD}, 2019.

\bibitem{Liu2018}
G.~Liu and O.~Schulte, ``Deep reinforcement learning in ice hockey for
  context-aware player evaluation,'' in \emph{International Joint Conference on
  Artificial Intelligence}, 2018.

\bibitem{Liu2019}
G.~Liu, Y.~Luo, O.~Schulte, and T.~Kharra, ``Deep soccer analytics: learning an
  action-value function for evaluating soccer players,'' \emph{Data Mining and
  Knowledge Discovery}, vol.~34, no.~2, 2020.

\bibitem{Kharrat2017}
T.~Kharrat, J.~L. Peña, and I.~McHale, ``Plus-minus player ratings for
  soccer,'' \emph{European Journal of Operational Research}, vol. 283, no.~2,
  2017.

\bibitem{Ian2012}
I.~G. McHale, P.~A. Scarf, and D.~E. Folker, ``On the development of a soccer
  player performance rating system for the english premier league,''
  \emph{Interfaces}, vol.~42, no.~4, pp. 339--351, 2012.

\bibitem{Statsbomb}
STATSBOBM, ``New data, new statsbomb radars,''
  \url{https://statsbomb.com/2018/08/new-data-new-statsbomb-radars/}, 2018.

\bibitem{Rudd}
\BIBentryALTinterwordspacing
S.~Rudd, ``A framework for tactical analysis and individual offensive
  production assessment in soccer using markov chains,'' 2018. [Online].
  Available: \url{http://nessis.org/nessis11/rudd.pdf}
\BIBentrySTDinterwordspacing

\bibitem{GoldnerKeith2012}
G.~Keith, ``A markov model of football: Using stochastic processes to model a
  football drive,'' \emph{Journal of Quantitative Analysis in Sports}, vol.~8,
  no.~1, 2012.

\bibitem{xT}
\BIBentryALTinterwordspacing
K.~Sing, ``Introducing expected threat (xt) modelling team behaviour in
  possession to gain a deeper understanding of buildup play.'' 2018. [Online].
  Available: \url{https://karun.in/blog/expected-threat.html}
\BIBentrySTDinterwordspacing

\bibitem{cervone2014}
D.~Cervone, A.~Amour, L.~Bornn, and K.~Goldsberry, ``Pointwise: Predicting
  points and valuing decisions in real time with nba optical tracking data,''
  in \emph{MIT Sloan Sports Analytics Conference}, 2014.

\bibitem{Xiangyu2020}
X.~Sun, J.~Davis, O.~Schulte, and G.~Liu, ``Cracking the black box: Distilling
  deep sports analytics,'' in \emph{ACM KDD}, 2020.

\bibitem{Dick2019}
U.~Dick and U.~Brefeld, ``Learning to rate player positioning in soccer,''
  \emph{Big Data}, vol.~7, no.~1, 2019.

\bibitem{jeff2016}
J.~Donahue, L.~A. Hendricks, M.~Rohrbach, S.~Venugopalan, S.~Guadarrama,
  K.~Saenko, and T.~Darrell, ``Long-term recurrent convolutional networks for
  visual recognition and description,'' in \emph{IEEE Conference on Computer
  Vision and Pattern Recognition}, 2016.

\bibitem{Teng2019}
T.~Xie, Y.~Ma, and Y.~Wang, ``Optimal off-policy evaluation for reinforcement
  learning with marginalized importance sampling,'' \emph{CoRR}, 2019.

\bibitem{DBLP:journals/corr/abs-1802-03493}
M.~Farajtabar, Y.~Chow, and M.~Ghavamzadeh, ``More robust doubly robust
  off-policy evaluation,'' \emph{CoRR}, 2018.

\bibitem{DBLP:journals/corr/JiangL15}
J.~Nan and L.~Lihong, ``Doubly robust off-policy evaluation for reinforcement
  learning,'' \emph{CoRR}, 2015.

\bibitem{dr2016}
N.~Jiang and L.~Li, ``Doubly robust off-policy value evaluation for
  reinforcement learning,'' in \emph{International Conference on Machine
  Learning}, 2016.

\end{thebibliography}

\clearpage

\begin{appendices}


\section{Raw data description}
\label{data}
The data used to conduct our experiments are collected by a company called InStat. The dataset provides both events and tracking information of 104 European soccer matches in 2017-2018 season. The original InStat datasets (both events and tracking data) have relative coordinates originated from the right-down corner of the pitch from the attacking team's perspective: (0 to 105 at x-axis and 0 to 68 at y-axis). The attack direction is always set to be from left to right, regardless of the home or away teams. The original columns of the event dataset are as follows: action name (pass, shot, dribble, ball out, foul, clearance, assist, and events such as goal, offside, own goal, challenges, etc.), (x,y) coordinates of the start and end, action result (successful or not), zone id, body id, time second, player name, team name, opponents, and match id. Due to the confidentiality of the InStat dataset, we are not allowed to share data. However, the respective results of the proposed algorithms can be reproduced using publicly available event and tracking datasets such as Wyscout\footnote{\url{https://figshare.com/collections/Soccer\_match\_event\_dataset/4415000/5}}. We provide the public codes available online\footnote{\url{https://github.com/Peggy4444/soccer\_RL}}.

\section{Transformed data description}
\label{data-qppendix}
The possession input to all spatiotemporal models can be constructed with most of the publicly available soccer logs, such as Wyscout dataset\footnote{\url{https://figshare.com/collections/Soccer\_match\_event\_dataset/4415000/2}}. However, generating the defensive pressure feature requires the access to tracking data, which is missing in some of the datasets. 

Moreover, the input to the network (for filling the replay buffer) for OPE should be prepared in the format of Table~\ref{policy}.

\begin{table*}[h!]
 \caption{Input to policy network}
  \label{policy}
  \begin{tabular}{|p{1cm}|p{3cm}|p{1cm}|p{1cm}|p{2cm}|p{2cm}|p{2cm}|p{0.5cm}|p{2cm}|}
  \hline
   \textbf{action}& \textbf{all action probabilities} & \textbf{episode}& \textbf{reward} & \textbf{possession number} & \textbf{possession team} &\textbf{possession feature 1 (10 dimensional)} & \textbf{...}& \textbf{possession feature n (10 dimensional)} \\
    \hline
   out & [0.36,0.25,0.14,0.25] & 1& -0.01& 1 & home   & [0.40,...,0.25] & ... & [2,...,3] \\
   \hline
   foul & [0.24,0.42,0.26,0.10] & 1 & 0.05 & 2 & home  & [0.37,...,0.65] & ... & [4,...,1] \\
   \hline
   shot & [0.22,0.25,0.33,0.20] & 1 &  0.28 & 3 & home & [0.48,...,0,32] & ... & [1,...,3] \\
   \hline
   . & . & . & . & . & .& . & ... & .\\
\hline
 \end{tabular}
\end{table*}

\section{Expected goal model with logistic regression}
\label{xg}
According to the state-of-the-art expected goal (xG) models in soccer analytics, we collected 15,225 shots, and labeled them by two classes of goal and not goal. Then we used logistic regression to estimate the probability of goal scoring, given the features of the shot: $P(goal|shot, X)$. For the implementation of logistic regression, we used scikit-learn Python package\footnote{\url{https://scikit-learn.org/stable/modules/generated/sklearn.linear\_model.LogisticRegression.html}}. This classification model showed the AUC of 80\% with 5-fold cross validation in the shot dataset.   

\section{Spatiotemporal models implementation}
\label{gpu}
\begin{itemize}
\item The spatiotemporal models are implemented using Keras sequential model.
\item Training the policy gradient method is performed via Keras with Tensorflow backend.
\item We conducted training and experimental results using a Tesla K80 GPU.
\end{itemize}

\section{Bayesian formula for deriving possession value (PV) }
\label{bayes}
\begin{gather*} 
PV(s) = Pr(shot | X)  Pr(goal | shot , X) \\
=\frac{P(shot, X)}{P(X)} \frac{P(goal, shot, X)}{P(shot, X) } \\
=\frac{P(goal, shot, X)}{P(X) } \\
=P(goal, shot | X) 
\end{gather*}

\section{Deriving gradients from off-policy policy gradient method}
\label{pg-off}

Given that training samples $x=\tau|\theta$ are sampled from behavioral policy $q(x)$, we can rewrite gradient as follows:

\begin{gather*} 
\nabla_\theta J(\theta)= \nabla_\theta E_x[\sum_x f(x) p(x)] \\
= E_x [\sum_x (f(x) \nabla_\theta p(x) + p(x) \nabla_\theta f(x))] \Leftarrow \mbox{Derivative product rule} \\
\approx E_x [\sum_x f(x) \nabla_\theta p(x)] \\
=E_x[\sum_x q(x) \frac{p(x)}{q(x)} f(x) \frac{\nabla_\theta p(x)}{p(x)}]\\
=E_x [\frac{p(x)}{q(x)} f(x) \nabla_\theta \log p(x)]\\
\end{gather*}

\section{off-policy policy evaluation with importance sampling}
\label{imp}
\begin{figure*}
  \includegraphics[width=\textwidth]{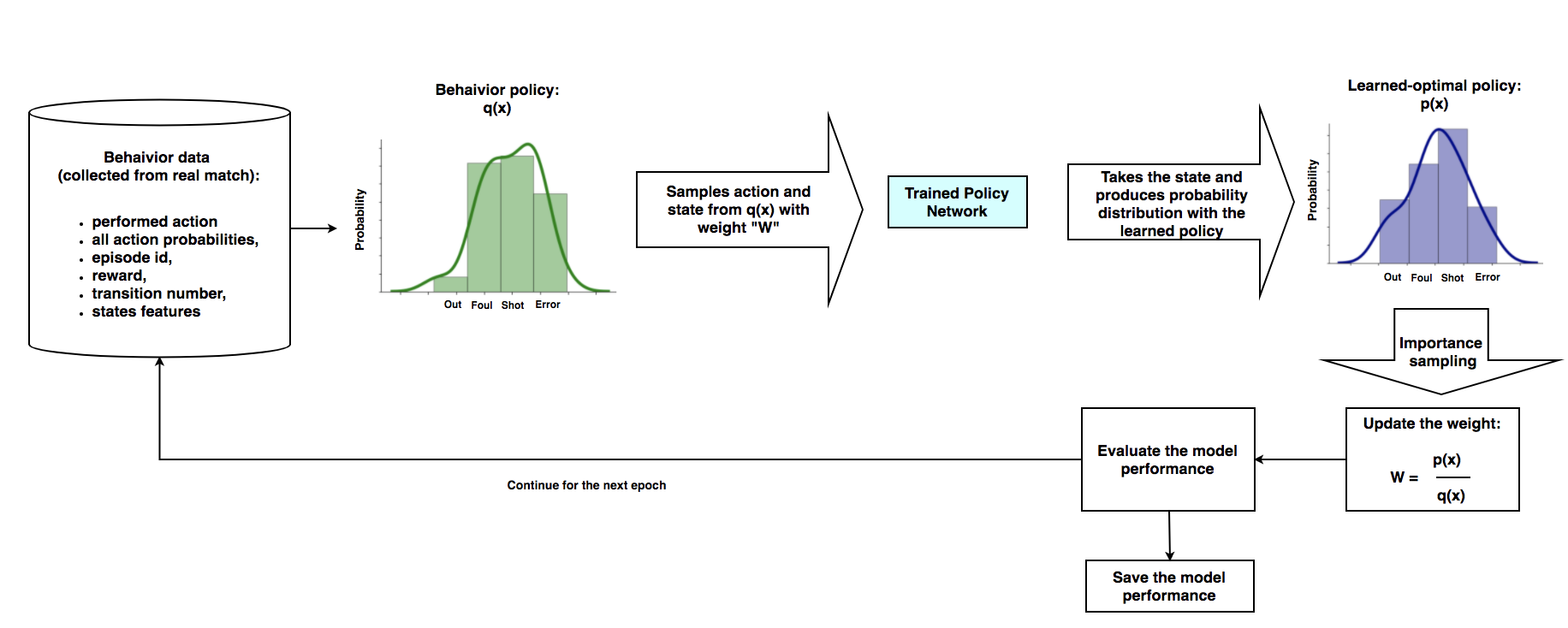}
  \caption{off-policy policy evaluation of the learned policy, based on the dataset from real soccer matches with importance sampling}
  \label{sampling}
\end{figure*}

The Importance Sampling method takes samples from behavioral policy $q(x)$ to evaluate the performance of target policy $p(x)$. Figure~\ref{sampling} shows the workflow of evaluation with this method.

\section{off-policy policy evaluation with doubly robust}
\label{dr}

According to the model by \cite{dr2016}, for a H-step trajectory $\tau=(s_1,a_1,r_1,\dots,s_H,a_H,r_H,s_{H+1})$ and $a_t\sim\pi$, we define state value function as $V^{\pi,H}(S)$ and action value function as $Q^{\pi,H}(S,A)$. Then if we are supplied with $\hat{Q}$, which is an estimate of action value function, we can apply doubly robust evaluator at each time-step $(t)$ as follows:

\begin{equation}
V_{DR}^{H+1-t}= \hat{V}(s_t) + \rho_t(r_t + \gamma V_{DR}^{H-t} -\hat{Q}(s_t,a_t)),
\end{equation}

where $\rho_t = \frac{\pi_1 (optimal)}{\pi_0(behavioral)}$, and $\hat{V}(s)= \sum_a \pi_1 (a|s)\hat{R}(s,a)$. Therefore, the doubly robust of the target policy value is $V_{DR}=V_{DR}^H$.

\section{Numerical Experiment Details}
\label{gamma}
In all the experiments we used the following parameters:
\begin{itemize}
\item discount factor $\gamma =0.99$
\item learning rate in policy gradient $\alpha =1e-4$
\item learning rate in deep learning $=0.01$
\end{itemize}
\end{appendices}

\end{document}